\documentclass[11pt, a4paper]{article}

\usepackage{geometry}
\usepackage{hyperref}
\usepackage{fancyhdr}
\usepackage{amsmath}
\usepackage{enumitem} % For better list control
\usepackage{tikz}
\usetikzlibrary{positioning, arrows.meta, fit, backgrounds, calc}
\usepackage[round]{natbib}
\hypersetup{
    colorlinks=true,
    linkcolor=black,
    citecolor=black,
    urlcolor=blue,
}

\begin{document}

\title{
    \textbf{LLM-based Triplet Extraction from Financial Reports}
}
\author{
    Dante Wesslund, Ville Stenström, Pontus Linde, and Alexander Holmberg \\[6pt]
    KTH Royal Institute of Technology
}
\date{February 2026}
\maketitle

\begin{abstract}
Corporate financial reports are a valuable source of structured knowledge for Knowledge Graph construction, but the lack of annotated ground truth in this domain makes evaluation difficult. We present a semi-automated pipeline for Subject-Predicate-Object triplet extraction that uses ontology-driven proxy metrics, specifically Ontology Conformance and Faithfulness, instead of ground-truth-based evaluation. We compare a static, manually engineered ontology against a fully automated, document-specific ontology induction approach across different LLMs and two corporate annual reports. The automatically induced ontology achieves 100\% schema conformance in all configurations, eliminating the ontology drift observed with the manual approach. We also propose a hybrid verification strategy that combines regex matching with an LLM-as-a-judge check, reducing apparent subject hallucination rates from 65.2\% to 1.6\% by filtering false positives caused by coreference resolution. Finally, we identify a systematic asymmetry between subject and object hallucinations, which we attribute to passive constructions and omitted agents in financial prose.
\end{abstract}

\section{Introduction}
In the data-driven landscape of modern finance, the ability to rapidly synthesize vast amounts of unstructured information into actionable insights is a critical competitive advantage. Corporate annual reports represent a particularly valuable source of such data. These documents are dense and highly heterogeneous, making manual extraction of structured knowledge both expensive and unscalable. Furthermore, financial disclosures are frequently characterized by complex syntax and heavy use of passive voice, which regulatory bodies such as the SEC have identified as a barrier to clarity in disclosure documents \citep{united1998plain}. Knowledge Graphs (KGs) offer a promising solution by organizing this unstructured data into structured descriptions of entities, relationships, and events \citep{KnowledgeGraphs}. 

While Large Language Models (LLMs) have demonstrated remarkable capabilities in natural language understanding and information extraction, leveraging them to automate KG construction introduces significant challenges. The primary obstacle is trust. LLMs are prone to hallucinations, i.e., generating content that is unfaithful to the source text \citep{HallucinationSurvey}. While some domains tolerate stochastic generation, the financial domain requires strict factual adherence; consequently, hallucinations pose an unacceptable risk, necessitating robust verification for any automated extraction process.

The standard approach to verifying LLM output is to rely on large, manually annotated ground truth datasets to calculate precision and recall. However, in dynamic environments, creating and maintaining these datasets is often infeasible. This work addresses this critical gap by developing a semi-automated pipeline for Subject-Predicate-Object (SPO) triplet extraction from corporate reports that does not rely on pre-annotated ground truth for primary evaluation.

Building on the framework proposed by \citet{Text2KGBench}, our pipeline utilises ontology-driven metrics to assess performance. We focus on two key dimensions of extraction quality: Ontology Conformance (OC), ensuring extracted triplets adhere to a predefined financial domain structure, and Faithfulness, ensuring the extracted information is grounded strictly in the source text without hallucination.

The main contributions of this work are:
\begin{itemize}
    \item[\textbf{C1:}] A comparative analysis of a manual pipeline using a static ontology versus a fully automated ontology induction strategy that generates ad-hoc schemas to assess report-specific specificity.
    \item[\textbf{C2:}] An evaluation of LLM performance under strict schema constraints, specifically identifying how different architectures influence ontology drift and adherence.
    \item[\textbf{C3:}] A presentation of a hybrid verification strategy (Regex + LLM-as-a-judge) that enables ground truth independent evaluation of faithfulness by filtering false positives from naive string matching.
\end{itemize}

Building upon these contributions, this study seeks to answer the following research questions: 
\begin{enumerate}
    \item[\textbf{RQ1:}] How does a dynamic, automatically induced ontology compare to a static, manually engineered ontology in terms of schema adherence and extraction faithfulness? 
    \item[\textbf{RQ2:}] To what extent does the choice of Large Language Model architecture influence the rate of relation hallucinations and ontology drift in financial contexts? 
    \item[\textbf{RQ3:}] Can a hybrid verification strategy effectively mitigate the high false-positive rates associated with strict regular expression matching?
\end{enumerate}

\section{Related Work}
The task of extracting structured knowledge from unstructured text has evolved significantly, moving from rigid, fully supervised pipelines to flexible, LLM-driven approaches. Traditionally, Triplet Extraction (TE) has been treated as a fully supervised learning problem. Early state-of-the-art models relied on end-to-end architectures, such as bidirectional Recurrent Neural Networks (RNNs) or, more recently, Transformer-based models, to jointly perform Named Entity Recognition (NER) and Relation Extraction (RE). These approaches rely heavily on large, high-quality labeled datasets such as the NYT distant-supervision corpus \citep{NYT}, WebNLG \citep{WebNLG}, and DocRED \citep{DocRED}. While effective on established benchmarks, their reliance on extensive ground truth data restricts their generality in dynamic domains like corporate financial reporting, where creating and managing such tailored datasets is often prohibitively expensive.

LLMs have shifted this toward paradigms that require less task-specific training data. Current approaches generally fall into two categories: fine-tuning and in-context learning. Recent work demonstrates that relatively smaller LLMs (e.g., 7B parameters) can be fine-tuned to outperform larger general-purpose models on specific extraction tasks, provided high-quality training data is available \citep{zhang2024fine}. Further, research has increasingly focused on zero-shot and few-shot prompting. While promising, standard LLMs often struggle in these settings without additional support, particularly when dealing with extensive documents. \citet{liu2024lost} identified a "lost-in-the-middle" phenomenon, observing that model performance degrades significantly when relevant information is located in the middle of long contexts. Addressing such limitations, \citet{papaluca2024zero} demonstrated that augmenting prompts with context dynamically retrieved from an external Knowledge Base (KB) significantly improves triplet extraction performance and makes LLMs competitive with strong supervised baselines compared to raw zero-shot prompting.

Beyond extracting triplets for a \textit{given} ontology, a parallel line of research investigates using LLMs for the \textit{creation} of the ontology itself, a process often referred to as Ontology Induction. This task is a critical sub-discipline of Knowledge Graph Engineering (KGE) \citep{KnowledgeGraphs}. However, traditional KGE frameworks typically focus on the curation of static, high-quality schemas meant for long-term use \citep{Meyer2023DevelopingScalableBenchmark}. There remains a gap in applying these induction techniques dynamically to handle the heterogeneity of single-shot financial reports, a challenge this study aims to address.

A critical limitation of deploying generative LLMs for knowledge extraction is their tendency to hallucinate. Standard evaluation of TE systems relies on comparing outputs to a ground truth using established metrics like precision and recall. In many real-world scenarios, such a ground truth does not exist. To address this gap, recent benchmarking frameworks have moved toward ontology-driven evaluation. The Text2KGBench \citep{Text2KGBench} framework specifically addresses this by measuring fact extraction not just against ground truth, but by Ontology Conformance (ensuring triplets use valid relations from a predefined ontology) and Faithfulness (verifying entities actually exist in the source text). To operationalize such verification without manual annotation, recent research established the "LLM-as-a-judge" paradigm \citep{zheng2023judging}, demonstrating that strong LLMs can effectively approximate human judgement in evaluating generated content.

\section{Background}
This section formalizes the task of ontology-driven knowledge extraction and defines the specific notation and metrics used to evaluate our pipeline without ground truth data.

\subsection{Problem description}
The objective of this work is to transform unstructured financial text into a structured Knowledge Graph. This involves the extraction of semantic triplets, defined as a set of subject, predicate, and object. These represent essential facts according to a predefined ontology. Corporate financial reporting lacks the standardised, manually annotated datasets required for traditional supervised evaluation. Consequently, the extraction quality is not measured against a ground-truth graph. Instead, performance is defined by two proxy objectives: adherence to the ontology and strict grounding in the source text.

Formally, we define the task of Ontology-Driven Triplet Extraction as finding a mapping function $f$ that takes an unstructured document corpus $\mathcal{D}$ and a target ontology $\mathcal{O}$ as input, and produces a Knowledge Graph $\mathcal{G}$ as output:

$$f(\mathcal{D}, \mathcal{O}) \rightarrow \mathcal{G}$$

where $\mathcal{G}$ is a set of semantic triplets.

In a standard supervised setting, the objective is to learn $f$ by minimizing a loss function against a ground truth graph $\mathcal{G}_{\text{gold}}$. In our setting, $\mathcal{G}_{\text{gold}}$ is unavailable. Therefore, the problem shifts to designing and evaluating $f$ based on two proxy objectives:

\begin{itemize}
    \item \textbf{Conformance:} $\mathcal{G}$ must only contain relations defined in $\mathcal{O}$.
    \item \textbf{Faithfulness:} Every triplet in $\mathcal{G}$ must be explicitly grounded in the text of $\mathcal{D}$.
\end{itemize}

These proxy objectives are operationalized in Section 3.3 by Ontology Conformance (OC), which quantifies Conformance, and the hallucination metrics SH, OH, and RH, which quantify the Faithfulness of the extracted triplets to the source text.

\subsection{Notation and Definitions}
We formalize the textual definitions provided by the Text2KGBench framework \citep{Text2KGBench} to define our components and evaluation metrics.

\textbf{Ontology ($\mathcal{O}$):} We define the ontology as $\mathcal{O} = (\mathcal{C}, \mathcal{R})$, where $\mathcal{C}$ is a set of concepts and $\mathcal{R}$ is a set of canonical relations. 

\textbf{Input Text ($\mathcal{D}$ and $c_i$):} 
The full corporate report is denoted as $\mathcal{D}$. This report is pre-processed into a sequence of smaller text chunks $\mathcal{D}=\{c_1, c_2, \dots, c_n\}$.

\textbf{Triplet ($t$):} A single extracted fact is represented as a triplet $t = (s,p,o)$, where $s$ is the subject entity, $p$ is the predicate (relation), and $o$ is the object entity. For example, given source $``\texttt{EBIT margin was 3.4 (4.9)\%}''$, a triplet $(\texttt{EBIT\_margin}, \texttt{has\_value}, \texttt{3.4\%})$ can be extracted.

\textbf{Extraction Output ($\mathcal{T}_i$):} For a given input chunk $c_i$, the pipeline produces a set of triplets $\mathcal{T}_i = \{t_{i,1}, t_{i,2}, \dots, t_{i,m} \}$. 

\textbf{Knowledge Graph ($\mathcal{G}$):} The final output defined in 3.1 is the union of all chunk outputs, $\mathcal{G}=\cup_{i=1}^n \mathcal{T}_i$.

\subsection{Evaluation Metrics}
We now define the metrics used to evaluate the mapping $f(\mathcal{D}, \mathcal{O})$ in the absence of a ground truth graph $\mathcal{G}_\text{gold}$. To evaluate extraction quality without ground truth, we utilise the following metrics that measure adherence to the inputs $c_i$ and $\mathcal{O}$:
\begin{itemize}
    \item \textbf{Ontology Conformance (OC):} Directly measures the Conformance objective from Section 3.1 by quantifying the percentage of triplets in a chunk output $\mathcal{T}_i$ whose predicate $p$ is a valid relation in $\mathcal{R}$.
    $$ OC(\mathcal{T}_i)=\frac{|\{(s,p,o) \in \mathcal{T}_i | p \in \mathcal{R} \}|}{|\mathcal{T}_i|}$$
    \item \textbf{Hallucination Metrics:} These metrics operationalize the Faithfulness objective from Section 3.1 by verifying whether each component of a triplet can be grounded in the source chunk $c_i$. To illustrate, we consider the source chunk
$c_i=$ ``\texttt{Net cash was SEK 27.1 (27.5) bn, which was largely driven by investing activities.''} and relation set
    $\mathcal{R} = \{ \texttt{reports\_metric}, \texttt{has\_value} \}$
    \begin{itemize}
        \item \textbf{Subject Hallucination (SH)}: A triplet $t=(s,p,o)$ has a subject hallucination if $s$ cannot be matched in the source chunk $c_i$. \\
        \emph{Ex:} $(\texttt{The Group}, \texttt{reports\_metric}, \texttt{Net cash})$. The subject \texttt{The Group} is inferred but is not present in $c$. 
        
        \item \textbf{Object Hallucination (OH):} A triplet $t=(s,p,o)$ has an object hallucination if $o$ cannot be matched in the source chunk $c_i$. \\
        \emph{Ex:} $(\texttt{Net cash}, \texttt{has\_value}, \texttt{SEK 27.2 bn}$. The amount (\texttt{SEK 27.2 bn}) is hallucinated (incorrect number) and not present in $c$.

        \item \textbf{Relation Hallucination (RH)}: A relation is considered hallucinated if it is not a member of the predefined ontology set $\mathcal{R}$. \\
        \emph{Ex:} $(\texttt{Net cash}, \texttt{driven\_by}, \texttt{investing activities})$. While \texttt{driven by} appears in the text $c$, strictly speaking $\texttt{driven\_by} \notin \mathcal{R}$, so it is flagged as a hallucination.
    \end{itemize}
\end{itemize}

\section{Methodology and Experimental Design}
This study employs a comparative analysis to evaluate LLM-based triplet extraction under different pipeline configurations. The methodology is structured around three core areas: (1) the constant data processing and extraction pipeline, (2) the experimental variables (models, corpora, and ontologies) that define our test matrix, and (3) the evaluation framework used to measure performance. 

\subsection{Core Extraction Pipeline}
While key components are varied for the experiment, the fundamental data processing and prompting architecture is held constant.

\subsubsection{Data Preprocessing and Segmentation}
Corporate annual reports are characterized by extreme information density. Empirical observations indicated that processing entire pages simultaneously often exceeded the effective attention capabilities of standard LLMs, leading to the exclusion of critical facts regardless of the theoretical context window size. This behavior aligns with the ``lost-in-the-middle'' phenomenon described by \citet{liu2024lost}, where performance degrades significantly when relevant information is located in the middle of long contexts.

To mitigate this, raw documents $\mathcal{D}$ are segmented into a sequence of chunks, $\mathcal{D}=\{c_1, c_2,\dots,c_n \}$. Based on early empirical tests, a chunk size of 5 sentences with an overlap of 0 was selected. This specific configuration was chosen to balance context sufficiency with extraction precision: the 5-sentence window ensures sufficient local context to resolve immediate co-references without introducing the noise associated with larger contexts. Furthermore, the zero-overlap constraint was strictly enforced to prevent the generation of redundant triplets across boundaries, thereby simplifying the aggregation step by ensuring each unique fact is presented to the model exactly once.

\subsubsection{LLM Prompting Strategy}
For each chunk $c_i$, a prompt is dynamically constructed. This prompt provides the LLM with three inputs: the text chunk $c_i$, the complete definition of the governing ontology $\mathcal{O}$ (defined in Section 4.2.3), and some example chunks with correct extraction (these are generic and not specific to the input chunk $c_i$). The prompt instructs the model to extract all (subject, predicate, object) triplets that strictly adhere to the provided ontology, outputting the results in a structured JSON format.

\subsection{Experimental Variables}
To test the trade-offs between generalization and specificity, we use three concepts: the model $\mathcal{M}$, the test report $\mathcal{D}$, and the ontology strategy $\mathcal{O}$. 

\subsubsection{Models ($\mathcal{M}$)}
Two distinct LLMs were selected for the extraction task:
\begin{itemize}
    \item $\mathcal{M}_1$: \textit{Gemini-2.5 Flash}
    \item $\mathcal{M}_2$: \textit{Llama 4 Maverick}
\end{itemize}

\subsubsection{Test Report ($\mathcal{D}$)}
The pipelines were evaluated against corporate annual reports from two distinct companies to test in-domain and out-of-domain performance. The source documents are in $\texttt{.txt}$ format and utilise Markdown to represent structural elements like headers and tables. The content consists of a mix of financial tables and descriptive narrative.
\begin{itemize}
    \item $\mathcal{D}_\text{Volvo}$: The full \textit{Volvo} 2024 annual report.
    \item $\mathcal{D}_\text{Elekta}$: The \textit{Elekta} 2022/2023 annual report. To manage computational resource constraints, the evaluation on this corpus was restricted to the first 25\% of the document text.
\end{itemize}

\subsubsection{Ontology Strategies ($\mathcal{O}$)}
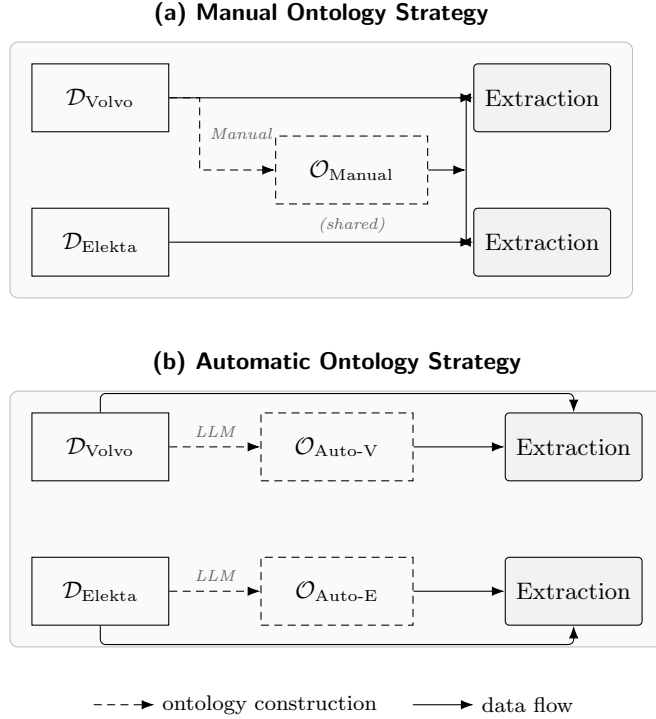
\begin{figure}[t]
    \centering
    \tikzset{
        >=Latex,
        line width=0.4pt,
        node distance=8mm and 12mm,
        doc/.style={draw, rectangle, minimum width=18mm, minimum height=9mm, align=center, font=\footnotesize},
        ont/.style={draw, densely dashed, rectangle, minimum width=20mm, minimum height=9mm, align=center, font=\footnotesize},
        proc/.style={draw, rectangle, rounded corners=1.5pt, minimum width=18mm, minimum height=9mm, align=center, font=\footnotesize, fill=black!5},
        data/.style={->, line width=0.4pt},
        construct/.style={->, densely dashed, line width=0.4pt},
        note/.style={font=\tiny\itshape, text=black!60}
    }
    \begin{tikzpicture}
    %% (a) MANUAL ONTOLOGY STRATEGY
    \node[doc] (d1) {$\mathcal{D}_{\text{Volvo}}$};
    \node[doc, below=10mm of d1] (d2) {$\mathcal{D}_{\text{Elekta}}$};
    \node[ont] at ($(d1.east)!0.5!(d2.east) + (24mm,0)$) (ontM) {$\mathcal{O}_{\text{Manual}}$};
    \node[proc, right=16mm of ontM |- d1] (e1) {Extraction};
    \node[proc, right=16mm of ontM |- d2] (e2) {Extraction};
    \draw[construct] (d1.east) -- ++(4mm,0) |- (ontM.west)
        node[pos=0.25, right, note] {Manual};
    \draw[data] (d1) -- (e1);
    \draw[data] (d2) -- (e2);
    \coordinate (ont-split) at ($(ontM.east) + (5mm, 0)$);
    \draw[data] (ontM.east) -- (ont-split);
    \draw[data] (ont-split) -- (ont-split |- e1) -- ([xshift=-2mm]e1.west);
    \draw[data] (ont-split) -- (ont-split |- e2) -- ([xshift=-2mm]e2.west);
    \node[note, below=1pt of ontM.south] {(shared)};
    \begin{scope}[on background layer]
        \node[draw=black!25, rounded corners=3pt, fill=black!2, inner sep=8pt, fit=(d1)(d2)(ontM)(e1)(e2)] (boxA) {};
    \end{scope}
    \node[font=\footnotesize\bfseries\sffamily, above=2pt of boxA.north] (titleA) {(a) Manual Ontology Strategy};
    %% (b) AUTOMATIC ONTOLOGY STRATEGY
    \node[doc, below=18mm of d2] (dA1) {$\mathcal{D}_{\text{Volvo}}$};
    \node[ont, right=of dA1] (oA1) {$\mathcal{O}_{\text{Auto-V}}$};
    \node[proc, right=of oA1] (eA1) {Extraction};
    \node[doc, below=10mm of dA1] (dA2) {$\mathcal{D}_{\text{Elekta}}$};
    \node[ont, right=of dA2] (oA2) {$\mathcal{O}_{\text{Auto-E}}$};
    \node[proc, right=of oA2] (eA2) {Extraction};
    \draw[construct] (dA1) -- (oA1) node[midway, above, note] {LLM};
    \draw[construct] (dA2) -- (oA2) node[midway, above, note] {LLM};
    \draw[data] (oA1) -- (eA1);
    \draw[data] (oA2) -- (eA2);
    \draw[data, rounded corners=2pt] (dA1.north) -- ++(0,2.5mm) -| (eA1.north);
    \draw[data, rounded corners=2pt] (dA2.south) -- ++(0,-2.5mm) -| (eA2.south);
    \begin{scope}[on background layer]
        \node[draw=black!25, rounded corners=3pt, fill=black!2, inner sep=8pt, fit=(dA1)(dA2)(oA1)(oA2)(eA1)(eA2)] (boxB) {};
    \end{scope}
    \node[font=\footnotesize\bfseries\sffamily, above=2pt of boxB.north] (titleB) {(b) Automatic Ontology Strategy};
    %% LEGEND
    \node[anchor=north, below=4mm of boxB.south, font=\scriptsize] {%
        \begin{tikzpicture}[baseline=-0.5ex, >=Latex]
            \draw[densely dashed, ->, line width=0.4pt] (0,0) -- (8mm,0) node[right, black] {ontology construction};
            \draw[->, line width=0.4pt] (42mm,0) -- (50mm,0) node[right, black] {data flow};
        \end{tikzpicture}%
    };
    \end{tikzpicture}
    \caption{Comparison of ontology strategies. \textbf{(a)}~Manual strategy: a single static ontology is derived from the Volvo report and applied to both documents, testing domain generalization. \textbf{(b)}~Automatic strategy: an LLM induces a unique ontology for each document before extraction, testing document-specific adaptation.}
    \label{fig:ontology_strategies}
\end{figure}

We compare two distinct approaches for defining the governing ontology $\mathcal{O}$, as illustrated in Figure \ref{fig:ontology_strategies}. While the manual approach follows traditional KGE practices, the automatic approach is motivated by the need to reduce human effort and eliminate semantic overlaps found in the relations of the manual ontology. 

\begin{enumerate}
 \item \textbf{Manual Ontology ($\mathcal{O}_\text{Manual}$)} This strategy tests for domain generalization. A static domain ontology was manually engineered. This ontology's concepts $\mathcal{C}$ and relations $\mathcal{R}$ were derived solely from an analysis of the $\mathcal{D}_\text{Volvo}$ report. This \textit{same} fixed ontology was then used for extraction on both $\mathcal{D}_\text{Volvo}$ (testing in-domain performance) and $\mathcal{D}_\text{Elekta}$ (testing out-of-domain generalization).
 \item \textbf{Automatic Ontology ($\mathcal{O}_\text{Auto}$)} This strategy tests a fully automated, report-specific ontology construction pipeline. At the start of processing each report, the ontology is empty. The document is then processed chunk by chunk. For each chunk $c_i$, \textit{Gemini-2.5 Flash} is prompted with the current ontology and asked only to propose additional entity types and relation labels that are required to describe the information in that chunk. The accepted additions are merged into the ontology and passed on to the next chunk. Running this procedure independently on each report yields two report-specific ontologies:
 \begin{itemize}
     \item $\mathcal{O}_\text{Auto-Volvo}$: Generated from $\mathcal{D}_\text{Volvo}$ and used \textit{only} for extraction on $\mathcal{D}_\text{Volvo}$
     \item $\mathcal{O}_\text{Auto-Elekta}$: Generated from $\mathcal{D}_\text{Elekta}$ and used \textit{only} for extraction on $\mathcal{D}_\text{Elekta}$
 \end{itemize}
\end{enumerate}

\subsection{Evaluation Framework}
The quantitative evaluation of each experimental run is based on the \textbf{Ontology Conformance (OC)} and \textbf{Hallucination (SH, OH, RH)} metrics from Section 3.3.

While OC and RH are calculated by deterministically checking the predicate $p$ against the given ontology's relation set $\mathcal{R}$, the faithfulness metrics (SH and OH) require a strategy to ground entities in the source text. To perform this verification at scale without a ground truth dataset, we compare two computational strategies:

\begin{itemize}
    \item \textbf{Baseline Verification:} A strict regular expression search is performed to locate the exact entity string verbatim within the source chunk $c_i$. This method serves as a baseline but is prone to high false positives from morphological variations.
    \item \textbf{Hybrid Verification:} First, the regex check is performed. If the entity is found verbatim, it is classified as "faithful" and the process stops. If the check fails, the entity is \textit{not} immediately flagged as a hallucination. Instead, it is passed to a secondary LLM-as-a-judge, following the evaluation paradigm established by \citet{zheng2023judging}. The judge is prompted with the source chunk $c_i$ and the extracted entity, and is tasked to determine if the entity is factually present in the text despite surface-level differences. 
\end{itemize}

For each experimental configuration (fixed choice of report $\mathcal{D}$, model $\mathcal{M}$, ontology strategy $\mathcal{O}$, and verification method), OC, SH, OH, and RH are reported as micro-averaged percentages over all extracted triplets. Concretely, the numerators and denominators in Section 3.3 are computed over the union of all triplets $\cup_i\mathcal{T}_i$ for that configuration, rather than averaged per chunk.

\section{Experimental Evaluation and Results}
This section presents the empirical results of our comparative analysis. We first validate the hybrid verification methodology, which is crucial for accurately measuring faithfulness without ground truth. We then present the primary quantitative results and provide a qualitative analysis of key findings.

\subsection{Validation of the Hybrid Verification Method}

\begin{table}[h] \centering 
\begin{tabular}{|l|c|c|c|c|} \hline \textbf{Verification Method} & \textbf{OC (↑)} & \textbf{SH (↓)} & \textbf{OH (↓)} & \textbf{RH (↓)} \\ 
\hline Baseline (Strict Regex) & 100 \% & 65.2 \% & 22.5 \% & 0 \% \\ \textbf{Hybrid (Regex + LLM-as-a-judge)} & 100 \% & \textbf{1.6 \%} & \textbf{0.24 \%} & 0 \% \\ \hline 
\end{tabular} 
\caption{Comparison of verification methods using report $\mathcal{D}_\text{Volvo}$, model $\mathcal{M}_1$, and ontology strategy $\mathcal{O}_\text{Manual}$. Arrows indicate desirable metric direction.} \label{tab:verification_comparison} 
\end{table}

To address RQ3, we first validate the hybrid verification methodology, which is essential for accurately measuring faithfulness without ground truth. Table \ref{tab:verification_comparison} presents a quantitative comparison between the baseline (strict regex) and the hybrid (Regex + LLM-as-a-judge) verification strategies.

Under the baseline method, the pipeline exhibited an SH rate of 65.2\% and an OH rate of 22.5\%. A manual audit of these "hallucinations" revealed that the vast majority were false positives, i.e., instances where the extraction model had correctly captured the semantic entity but altered the surface form.

The corrected metrics show a SH rate of only 1.6\% and an OH rate of 0.24\%. To validate the judge itself, a random sample of 10 fallback decisions was manually reviewed, confirming a 100\% accuracy rate.

\subsection{Comparative Analysis of Extraction Pipelines}
Having validated the measurement technique, we present the primary results of this study in Table \ref{tab:main_results} to address RQ1 regarding ontology strategies and RQ2 regarding model performance. The quantitative results reveal distinct performance patterns across the three experimental variables:
\begin{table} \centering
    \begin{tabular}{|l|l|l|c|c|c|c|}
        \hline \textbf{Report} & \textbf{Model} & \textbf{Ontology} & \textbf{OC (↑)} & \textbf{SH (↓)} & \textbf{OH (↓)} & \textbf{RH (↓)} \\
        \hline $\mathcal{D}_\text{Volvo}$ & $\mathcal{M}_1$ & $\mathcal{O}_\text{Manual}$ & 100 \% & 1.6 \% & 0.24 \% & 0 \% \\
        \hline $\mathcal{D}_\text{Volvo}$ & $\mathcal{M}_1$ & $\mathcal{O}_\text{Auto}$ & 100 \% & 0 \% & 0 \% & 0 \%\\
        \hline $\mathcal{D}_\text{Volvo}$ & $\mathcal{M}_2$ & $\mathcal{O}_\text{Manual}$ & 92.7 \% & 3.6 \% & 0.9 \% & 7.3 \%\\
        \hline $\mathcal{D}_\text{Volvo}$ & $\mathcal{M}_2$ & $\mathcal{O}_\text{Auto}$ & 100 \% & 1.2 \% & 0.54 \% & 0 \% \\
        \hline $\mathcal{D}_\text{Elekta}$ & $\mathcal{M}_1$ & $\mathcal{O}_\text{Manual}$ & 100 \% & 1.9 \% & 0.34 \% & 0 \% \\
        \hline $\mathcal{D}_\text{Elekta}$ & $\mathcal{M}_1$ & $\mathcal{O}_\text{Auto}$ & 100 \% & 0 \% & 0 \% & 0 \% \\
        \hline $\mathcal{D}_\text{Elekta}$ & $\mathcal{M}_2$ & $\mathcal{O}_\text{Manual}$ & 89.4 \% & 4.2 \% & 1.3 \% & 10.6 \%\\
        \hline $\mathcal{D}_\text{Elekta}$ & $\mathcal{M}_2$ & $\mathcal{O}_\text{Auto}$ & 100 \% & 1.5 \% & 0.28 \% & 0 \%\\
        \hline
    \end{tabular}
\caption{Aggregate performance metrics across all experimental configurations, measured using the hybrid verification method. Arrows indicate desirable metric direction.}
\label{tab:main_results}
\end{table}

\paragraph{Impact of Ontology Strategy} 
The $\mathcal{O}_\text{Auto}$ strategy consistently achieved 100\% Ontology Conformance (OC) across both models and reports. In contrast, the $\mathcal{O}_\text{Manual}$ strategy resulted in conformance degradation for model $\mathcal{M}_2$, particularly in the out-of-domain $\mathcal{D}_\text{Elekta}$ report (89.4\% OC).

\paragraph{Model Performance} 
Comparing the two models, $\mathcal{M}_1$ demonstrated superior adherence to strict constraints, maintaining 100\% OC and 0\% RH in all configurations. $\mathcal{M}_2$ exhibited a tendency toward Relation Hallucination (RH) when constrained by the manual ontology, generating predicates not present in $\mathcal{R}$ (7.3\% on Volvo, 10.6\% on Elekta). Furthermore, $\mathcal{M}_1$ achieved notably lower Subject Hallucination (SH) rates, achieving 0\% SH in both $\mathcal{O}_\text{Auto}$ configurations, whereas $\mathcal{M}_2$ persisted with minor hallucinations (1.2\%--1.5\%).

\paragraph{Corpus Effects} 
Across ontology strategies, $\mathcal{M}_1$ maintained 100\% OC and 0\% RH on both $\mathcal{D}_\text{Volvo}$ and $\mathcal{D}_\text{Elekta}$. Under $\mathcal{O}_\text{Manual}$, SH and OH were slightly higher on $\mathcal{D}_\text{Elekta}$ (1.9\% SH, 0.34\% OH) than on $\mathcal{D}_\text{Volvo}$ (1.6\% SH, 0.24\% OH), while both reports reached 0\% SH and 0\% OH under $\mathcal{O}_\text{Auto}$. For $\mathcal{M}_2$ with $\mathcal{O}_\text{Manual}$, OC decreased from 92.7\% on $\mathcal{D}_\text{Volvo}$ to 89.4\% on $\mathcal{D}_\text{Elekta}$, and RH increased from 7.3\% to 10.6\%; SH and OH also rose from 3.6\% to 4.2\% and from 0.9\% to 1.3\%, respectively. In contrast, with $\mathcal{O}_\text{Auto}$, $\mathcal{M}_2$ achieved 100\% OC and 0\% RH on both reports, with comparably low SH and OH rates (1.2\% / 0.54\% on $\mathcal{D}_\text{Volvo}$ and 1.5\% / 0.28\% on $\mathcal{D}_\text{Elekta}$).
\\ \\
Overall, these results show that ontology strategy, model choice, and domain shift all have measurable effects on schema adherence and hallucination rates under the hybrid verification method. The quantitative data allows for preliminary answers to the research questions. Regarding RQ3, the sharp decline in hallucination rates between the baseline and hybrid methods in Table \ref{tab:verification_comparison} suggests that semantic verification is necessary filter surface-level evaluation artefacts. For RQ1, the consistent $100\%$ OC of $\mathcal{O}_\text{Auto}$ across all configurations indicates that document-specific induction effectively eliminates schema violations. Finally, regarding RQ2, the performance gap between models under the manual strategy shows that $\mathcal{M}_1$ is more inherently conservative, while $\mathcal{M}_2$ is more susceptible to ontology drift when the schema is not tailored to the specific report.

\section{Summary and discussion}
This section provides a qualitative analysis of the experimental findings to address the mechanisms underlying the research questions. Specifically, we explore the efficiency gains provided by document-specific ontology induction (RQ1), the architectural differences in schema adherence (RQ2), and the linguistic drivers of hallucination asymmetry (RQ3). We also discuss limitations and state our conclusions. 

\subsection{Performance of Automatic Ontology}
The experimental results address RQ1 by demonstrating a universal performance increase when utilizing the automatic ontology strategy ($\mathcal{O}_\text{Auto}$) compared to the static manual baseline ($\mathcal{O}_\text{Manual}$). As shown in Table \ref{tab:main_results}, performance metrics improved across all metrics for both reports ($\mathcal{D}_\text{Volvo}$, $\mathcal{D}_\text{Elekta}$) and both models ($\mathcal{M}_1$, $\mathcal{M}_2$). Notably, the Subject Hallucination (SH) and Object Hallucination (OH) rates were consistently minimized under $\mathcal{O}_\text{Auto}$, indicating a higher degree of faithfulness between the extracted triplets and the source contents within the report.

Similarly, we observed a reduction in Relation Hallucination (RH), even for model $\mathcal{M}_2$, which exhibited significant ontology drift under the manual strategy. We attribute this broad performance uplift to the dynamic nature of the automatic ontology, which is substantially more compact than the manual ontology. Unlike the manual approach, the automatic process utilises only the entities and relations deemed necessary for the specific chunk $c_i$ which results in a much smaller ontology overall. These results answer RQ1 by showing that document-specific induction is more robust than static human engineering for maintaining schema adherence.

This compactness confers a twofold advantage to the extraction process:
\begin{enumerate}
    \item \textbf{Reduction of Ambiguity:} By limiting the ontology to only relevant concepts, the search space for the LLM is reduced. There are fewer candidate relations and entities to choose from, and the distinctions between available options are sharper.
    \item \textbf{Context Window Efficiency:} A smaller ontology occupies less of the context window. This leaves a relatively larger proportion of the available context for the actual contents of the chunk, enabling the LLM to reason more effectively and perform more accurate extractions.
\end{enumerate}

\subsection{Model-Specific Adherence to Schema Constraints}
The quantitative results in Table \ref{tab:main_results} provide a definitive answer to RQ2 by revealing systematic differences in how the two models respond to schema constraints. Model $\mathcal{M}_1$ exhibits strictly conservative behaviour with respect to the ontology, maintaining 100\% Ontology Conformance (OC) and 0\% Relation Hallucination (RH) across all configurations, both for $\mathcal{O}_\text{Manual}$ and $\mathcal{O}_\text{Auto}$.

By contrast, when paired with the manual ontology, $\mathcal{M}_2$ occasionally proposes relations that are not licensed by the target schema, leading to non-zero RH and reduced OC on both reports. This behaviour is particularly pronounced on the out-of-domain configuration. In this setting, OC decreases and RH increases relative to the in-domain results. Consequently, $\mathcal{M}_2$ appears more inclined to introduce additional predicate types it considers reasonable, even when these are not part of the predefined relation set $\mathcal{R}$.

When both models are instead combined with the automatically induced ontology $\mathcal{O}_\text{Auto}$, the picture changes: OC returns to 100\% and RH drops to 0\% for $\mathcal{M}_2$, while $\mathcal{M}_1$ maintains its already strict adherence. This suggests that a schema which is adapted to the document and to the extraction behaviour of the model can substantially mitigate ontology drift, particularly for architectures that are otherwise prone to over-generating relation labels.

The performance disparity under the manual ontology strategy can be attributed to fundamental differences in instruction following capabilities regarding negative constraints. The manual strategy imposes a high cognitive load, requiring the model to treat the ontology as a strict closed-world system. $\mathcal{M}_1$ successfully interprets the ontology as a hard constraint, rejecting valid linguistic relationships (e.g., \texttt{driven by}) if they do not map to $\mathcal{R}$. Conversely, $\mathcal{M}_2$ prioritises semantic fluency over schematic rigidity; despite explicit instructions to the contrary, it frequently hallucinates relations that are linguistically natural but ontologically invalid. This indicates that while $\mathcal{M}_2$ effectively captures semantic meaning, it struggles to suppress its generative tendencies in favor of the strict constraint satisfaction required for controlled KG construction.

\subsection{Asymmetry in Hallucination Metrics (SH vs. OH)}
The experimental results provide a qualitative answer to RQ3 by revealing a pronounced asymmetry between Subject Hallucinations (SH) and Object Hallucinations (OH). As shown in Table \ref{tab:verification_comparison}, the strict regex baseline substantially overestimates hallucination rates, particularly for subjects, while the hybrid verification method corrects most of these evaluation artefacts.

The much larger reduction in SH than in OH under hybrid verification indicates that many baseline ``subject hallucinations'' arise from legitimate coreference resolution. The extraction pipeline often replaces implicit or pronominal subjects with an explicit entity name, which breaks exact string matching but remains semantically faithful when assessed by the semantic verification step.

Even after this correction, residual SH remain consistently higher than OH. This suggests that when hallucinations do occur, they are more likely to involve fabricated or mis-specified subjects than objects. A plausible explanation is tied to the linguistic properties of financial reporting: the frequent use of passive constructions and omitted agents forces the model to infer subjects that are not explicitly realised in the text. By contrast, objects are more often realised as concrete, localised spans such as numerical amounts, dates, and instrument names, which can be copied verbatim, leading to lower object hallucination rates.

Taken together, these observations indicate that subject positions constitute a structurally more challenging slot for faithful extraction than object positions. The findings confirm that evaluation protocols must be sensitive to coreference and implicit arguments to avoid conflating genuine hallucination with surface-form variation.

\subsection{Limitations}
This study has several limitations that qualify the interpretation of the results. First, the empirical analysis is based on only two English annual reports from large, publicly listed firms. The findings may therefore not generalise to other document genres, languages, or smaller organisations.

Second, for the out-of-domain corpus $\mathcal{D}_\text{Elekta}$, the evaluation covers only the first 25\% of the document due to computational constraints. Later sections may exhibit different disclosure patterns and entity types, so the reported error rates should be viewed as partial.

Third, the evaluation focuses on ontology conformance and hallucination rates (SH, OH, RH), but does not directly measure recall. Highly conservative extraction behaviour can thus appear favourable in terms of faithfulness while still omitting a substantial number of valid triplets.

Fourth, the hybrid verification strategy depends on an automatic semantic judge. A small manual audit supported its decisions on the sampled cases, but the overall reliability of the metrics remains partly contingent on the behaviour and potential biases of this verifier.

Fifth, a combined ontology utilizing both a stable core schema and a corpus-specific automatically generated extension was not explored in this work.

Sixth, we only manually reviewed a small sample of the total hybrid verifications. Although they were all correct, it is possible this ratio does not hold up for all extractions.

Finally, both ontology strategies introduce additional sources of bias. The manual ontology $\mathcal{O}_\text{Manual}$ was engineered from a single report and reflects specific modelling choices, while the automatic ontology $\mathcal{O}_\text{Auto}$ is induced from the same documents used for extraction and may overfit to document-specific phrasing.

\subsection{Conclusion}
This work proposed a methodology for extracting and verifying structured knowledge from financial texts without relying on annotated ground truth. 

The first contribution addresses RQ1 regarding the role of ontology design. The comparison between the manually engineered ontology $\mathcal{O}_\text{Manual}$ and the automatically induced ontology $\mathcal{O}_\text{Auto}$ suggests that adapting the schema to the document can substantially reduce schema violations for models that are otherwise prone to over-generating relation labels, while preserving high adherence for more conservative models.

The second contribution, addressing RQ2, is a comparative analysis of schema adherence across models and reports. The results indicate that strict conformity to a predefined ontology is strongly model-dependent: one model adheres almost perfectly to the schema, while the other exhibits ontology drift under the manual ontology, especially on out-of-domain text. Together with the observed asymmetry between subject and object hallucinations, this highlights that both model choice and slot type have a material impact on the faithfulness of extracted triplets.

The third contribution, addressing RQ3, is the validation of a hybrid verification strategy that combines deterministic pattern matching with a semantic verification step. The experiments show that exact string matching alone substantially overestimates hallucination rates, particularly for subjects in the presence of implicit arguments and passive constructions, whereas the hybrid procedure yields more reliable estimates of faithfulness.

Future work should investigate ontology strategies that combine a stable core schema with a document-specific extension, aiming to retain the interpretability of the manual ontology while capturing issuer- and sector-specific nuances. Further work should also extend the evaluation to a broader set of issuers and document types.

\bibliography{references}
\end{document}